\documentclass{nolta}
\usepackage[dvips]{graphicx}
\usepackage{url}

\usepackage{algorithm}
\usepackage{algpseudocode}
\usepackage{graphicx}
\usepackage{subcaption}

\Vol{13}%
\No{2}%
\Year{2022}%
\Month{4}%

\title{Set-based comprehensive learning and particle swarm optimization with memory
for discrete optimization problem}

\AUTHOR{%
\author{Yuta Chikuba}{1}, 
\author*{Takahiro Hino}{2}, 
\author[a]{Michiharu Maeda}{1}
}

\AFFILIATE{%
\affiliate{Department of Computer Science and Engineering, Fukuoka Institute of Technology\\
3-30-1 Wajirohigashi, Higashi-ku, Fukuoka, Fukuoka 811-0295, Japan}{1}
\affiliate{Central Japan Office, JFE Systems Inc. \\
1-1 Kawasaki-cho, Handa, Aichi 475-8611, Japan}{2}
}

\EMAIL{%
\email{maeda@fit.ac.jp}{a}
}

\received{10}{18}{2021}
\revised{12}{18}{2021}
\published{4}{1}{2022}

\begin{document}

\begin{abstract} 
 
  This paper describes a novel algorithm of set-based comprehensive learning and particle swarm optimization with memory for discrete optimization problem.
  Our algorithm is an approach of searching for the best position in a set by referring to the current position when updating the solution.  
  In order to show the validity of our algorithm, we examine numerical experiments compared to existing algorithms.
  
\end{abstract}

\begin{keywords}
 comprehensive learning, particle swarm optimization, discrete space, optimization problem
\end{keywords}

\maketitle

\section{Introduction}
For optimization problem, it is difficult to obtain an optimal solution and calculation requires a huge amount of time. 
In this situation, metaheuristics have been a focus attention for solving optimization problem [1].
Metaheuristics involve, for example, genetic algorithm (GA) [2] that imitates the evolutionary process of biology,
ant colony optimization (ACO) [3] that imitates the foraging behavior of ant, and
particle swarm optimization (PSO) [4] that imitates the simple social model such as birds flocking and fish schooling.
For optimization problem, there are continuous optimization problem in continuous space and discrete optimization problem in discrete space.
Typical examples of discrete optimization problem include traveling salesman problem (TSP) which searches for the shortest possible route that 
one visits each given city once and returns to the origin city.
Though the traditional PSO in which particles have a position and velocity is designed to work only in continuous space,
discrete particle swarm optimization (DPSO) [5] operates in binary space by using sigmoid function.
Memory binary particle swarm optimization (MBPSO) [6] based on DPSO decides a position according to the previous position and its mechanism is simple and effective.
Although DPSO updates a position of particle without considering the previous position, MBPSO memorizes a position which is updated according to a new defined velocity.
Set-based comprehensive learning particle swarm optimization (SCLPSO) [7] operates in discrete space, and all arithmetic operators in the position and velocity updating rules
in the original PSO are redefined by new operators and procedures. 
SCLPSO can solve discrete optimization problem with high quality and is successfully applied to the large-scale problems.
For knapsack problem, set-based particle swarm optimization with status memory has been suggested and shown significant results [11].
For an attempt to traveling salesman problem, set-based particle swarm optimization with memory state has been discussed [12].

In this paper, we present a novel algorithm of set-based comprehensive learning and particle swarm optimization with memory (SCLPSOM)
to search the best position in a set by referring the current position when updating the solution.
Experimental results are compared the proposed algorithm to simple genetic algorithm (SGA) [2], max-min ant system (MMAS) [8], DPSO, MBPSO and SCLPSO
for traveling salesman problem.

This paper is organized as follows. In the next section, DPSO, MBPSO, and SCLPSO are described.
Section 3 presents SCLPSOM. Section 4 describes numerical computation and  result. Finally, Section 5 summarizes conclusions.

\section{Preliminary}

\subsection{Particle Swarm Optimization}
Particle swarm optimization (PSO) [4] is an optimization approach that achieves the social model of birds flocking and fish schooling. 
Swarms move to a particle which finds out the best solution. Each particle has a position and velocity which are a candidate solution and a vector constructed the current position
and the best position, respectively. Each particle stores the base position previously encountered by itself and the best position of all particles.
The traditional PSO algorithm is only operated in continuance space. Updating rule of velocity and position of $i$th particle are defined as follows:
\begin{equation}
  v^{k+1}_{ij} = \omega v^{k}_{ij} + c_{1} r_{1} (P_{ij} - x^{k}_{ij}) + c_{2} r_{2} (G_{j} - x^{k}_{ij})
\end{equation}
\begin{equation}
  x^{k+1}_{ij} = x^{k}_{ij} + v^{k}_{ij}  \quad (1 \leq i \leq n, 1 \leq j \leq d)
\end{equation}
where $n$ is number of particles and $d$ is number of dimension of search space. $v^{k}_{ij}$ is $j$-th dimension of $i$-th particle's velocity of $k$-th generation and 
$x^{k}_{ij}$ is $j$-th dimension of $i$-th particle's position of $k$-th generation. $P_{ij}$ is $j$-th dimension of the best position of $i$-th particle and
$G_{j}$ is $j$-th dimension of the best position of particle swarm. $c_{1},c_{2}$ and $\omega$ are positive coefficients, $r_{1}$ and $r_{2}$ are random uniform values
in $[0.0,1.0]$. The procedure of the PSO algorithm is summarized as follows:

\begin{description}

  \item[Step1:] Initialize the positions and velocities.
  \item[Step2:] Evaluate the fitness on current position of each particle.
  \item[Step3:] Update the best solution of each particle.
  \item[Step4:] Update the best position of all particles.
  \item[Step5:] Update the velocity according Eq.(1).
  \item[Step6:] Update the position according Eq.(2).
  \item[Step7:] If the current iteration is smaller than the maximum number of iteration, then go to step2, otherwise terminate. 

\end{description}

\subsection{Discrete Particle Swarm Optimization}
Discrete particle swarm optimization (DPSO) [5] is designed to work particle swarm optimization in discrete space.
In DPSO, the position is defined as the bit string for discrete problem. The velocity is defined as the probability of real number.
The updating rule of position is denoted as follows:
\begin{equation}
  x^{k+1}_{ij} = 
  \left\{
  \begin{array}{ll}
    1 & ( \mathrm{rand}() < \mathrm{sig}(v^{k+1}_{ij})) \\
    0 & (\mathrm{otherwise})
  \end{array}
  \right.
\end{equation}
where $\mathrm{rand}()$ is a uniform random value in $[0.0,1.0]$. $\mathrm{sig}(*)$ denotes a sigmoid function:
\begin{equation}
  \mathrm{sig}(v^{k+1}_{ij}) =  \frac{1}{1 + e^{-v^{k+1}_{ij}}}
\end{equation}
\subsection{Memory Binary Particle Swarm Optimization}
Memory binary particle swarm optimization (MBPSO) [6] is a new binary PSO that stores a previous position and updates a position according to a redefined velocity . 
Thus, the updating rules of velocity and position are modified. 
The updating rules of velocity is defined as follows:
\begin{equation}
  v^{k+1}_{ij} = \omega v^{k}_{ij} + c_{1}r_{1}E(P_{ij},x^{k}_{ij}) + c_{2}r_{2}E(G_{j},x^{k}_{ij})
\end{equation}
where  $E(*,*)$ is the function that has a value of 1 if current status and target status are the same, and a value of -1 otherwise, as shown in the following equation.
\begin{equation}
  E(A,B) =
  \left\{
  \begin{array}{ll}
    1 & (A = B) \\
    -1 & (\mathrm{otherwise})
  \end{array}
  \right.
\end{equation}
The updating rules of position is defined as follows:
\begin{equation}
  x^{k+1}_{ij} =
  \left\{
  \begin{array}{ll}
    x^{k}_{ij} & (\mathrm{rand}() < \mathrm{sig}(v^{k}_{ij})) \\
    \bar{x}^{k}_{ij} & (\mathrm{otherwise})
  \end{array}
  \right.
\end{equation}
where velocity $v^{k}_{ij}$ represents the probability of $j$-th dimension of particle $i$ to retain the current state.
$\bar{x}^{k}_{ij}$ is bit inversion of $x^{k}_{ij}$. 

\subsection{Set-Based Particle Swarm Optimization}
Set-based particle swarm optimization(SPSO)[7] extends the application of PSO for solving combinatorial optimization problems in discrete space.
A velocity is defined as a set with possibility and a position is defined as the candidate solution for discrete optimization problems.
All arithmetic operators and procedures in the velocity and position updating rules are redefined by new operators and procedures.\\
In SPSO of TSP, a universal set $E$ and a candidate solution $X$ are defined in [7].
Each arc $(i,j) \in E$ between cities $i$ and $j$ are assigned length $d_{ij}$.
$E^{j}$ is j-th dimension of $E$. $E^{j}$ only includes the arcs that start from node $j$. 
$X^{j}$ is the subset of $E^{j}$ with two arcs that connects with $j$.

In SPSO, velocity is set with possibility defined as follows:
\begin{equation}
  V^{j}_{i} = \{ e / p(e) ~|~ e \in E^{j} \}
\end{equation}
where $V^{j}_{i}$ is the set with possibility of $j$-th dimension of $i$-th particle. 
$E^{j}$ is a universal set of $j$-th dimension. 
$V^{j}_{i}$ has a possibility $p(e) \in [0.0 , 1.0]$ of each element $e \in E$. 
SPSO is likely to select the same position as the current position. It generate solution similar to the current generation solution. 
If the swarm size is small, the algorithm is more likely to be trapped into local minimum.

To avoid this situation, for the velocity updating, we use comprehensive learning particle swarm optimization(CLPSO) [9].
The updating rule of velocity is denoted as follows:
\begin{equation}
  V^{j}_{i} = \omega V^{j}_{i} + cr^{j}(PBest^{j}_{f_{i}(j)} - X^{j}_{i})
\end{equation}
where $X^{j}_{i}$ is position of $j$-th dimension of $i$-th particle. The position is given by a crisp set.
$PBest^{j}_{f_{i}(j)}$ is the best solution of $j$-th dimension of $i$-th particle. 
$f_{i}(j)$ is given as the following description [9]. First, a uniform random value $R$ in $[0.0,1.0]$ is generated.
If $R$ is larger than a parameter with $i$-th particle , $f_{i}(j)$ is substituted into $i$ 
otherwise the algorithm applies position of particle of high fitness to two randomly selected particles.

In addition, all arithmetic operators in Eq.9 are redefined in the discrete space as follows.

The multiplication operator between a coefficient and a set with possibilities is defined as follows:
\begin{eqnarray} 
	\omega V &=& \{e / p^{\prime}(e) ~| ~e \in E\}\\
	p^{\prime}(e) &=& 
	\left\{
	\begin{array}{ll}
	1 & ( \omega \times p(e) > 1) \\
	\omega \times p(e) & (\mathrm{otherwise})
	\end{array}
	\right.
\end{eqnarray}
where $\omega$ is a coefficient, $V$ is a set with possibilities.

For example, if a coefficient and a set with possibilities are $\omega = 0.8$ and $V_{i} = \{(1,2)/0.2,(1,3)/0.7,$ $(3,5)/0.7,(4,5)/0.5\}$,
we obtain $\omega V_{i} = \{(1,2)/0.16,(1,3)/0.56,(3,5)/0.56,(4,5)/0.4\}$.

Given two crisp sets $A$ and $B$, the minus operator between two crisp sets is defined as follow:
\begin{equation}
	A - B = \{ e ~| ~e \in A  ~\mathrm{and}  ~e \notin B \}
\end{equation}		
For example, $PBest^{1}_{f_{i}(j)} = \{(1,2),(1,6)\}$ and $X^{1}_{i} = \{(1,3),(1,6)\}$ are given, we obtain $PBest^{1}_{f_{i}(j)} - X^{1}_{i} = \{(1,2)\}$.
$PBest^{j}_{f_{i}(j)} - X^{j}_{i}$ is to find out the elements used by the best position $PBest^{j}$ but not used by the current position $X^{j}$.
Such elements may have great potential to improve $X_{i}$.

The multiplication operator between a coefficient and a crisp set is defined as follow:
\begin{eqnarray}
	c E^{\prime} &=& \{e/p^{\prime \prime}(e) ~| ~e \in E\}\\
	p^{\prime \prime}(e) &=&
	\left\{
  	\begin{array}{ll}
			1 & (e \in E^{\prime} ~\mathrm{and} ~c > 1) \\
			c & (e \in E^{\prime} ~\mathrm{and} ~0 \leq c \leq 1)\\
			0 & (e \notin E^{\prime})
	  \end{array}
	\right.
\end{eqnarray}
where $E^{\prime}$ is a subset of a universal set $E$. $c~(c \geq 0)$ is a coefficient, $r^{j} \in [0.0, 1.0]$ is a uniform random number of $j$-th dimension.
For example, if $cr^{1}=0.7$, we obtain $cr^{1} (PBest^{1}_{f_{i}(j)} - X^{1}_{i}) = \{(1,2)/0.7\}$.

The plus operator between two sets with possibilities is defined as follow.
\begin{equation}
	V_{1} + V_{2} = \{ e / \mathrm{max}(p_{1}(e),p_{2}(e)) ~| ~e \in E) \}
\end{equation}
where $V_{1} = \{e/p_{1}(e) ~| ~e \in E \}$ and $V_{2} = \{e/p_{2}(e) ~| ~e \in E \}$ are defined on $E$.
$V_{1} + V_{2}$ is set to the larger one between $p_{1}(e)$ and $p_{2}(e)$. For example, $\omega V^{1} = \{(1,2)/0.16,(1,3)/0.56\}$ and 
$cr(PBest^{1}_{f_{i}(j)} - X^{1}) = \{(1,2)/0.7 \}$ are given, we obtain $V^{1}_{i} =  \omega V^{1} + cr^{1}(PBest^{1}_{f_{i}(j)} - X^{1}) = \{(1,2)/0.7,(1,3)/0.56\}$.

We define $K^{j}_{i}$ that keeps large possibility and converts a set with possibility to a crisp set.
$K^{j}_{i}$ of $j$-th dimension of $i$-th particle is defined as follows:
\begin{equation}
  K^{j}_{i} = \{ e~|~e / p(e) \in V^{j}_{i} ~\mathrm{and} ~ p(e) \geq \alpha_{i} \}
\end{equation}
where $\alpha_{i}$ is a random value $[0.0,1.0]$ of $i$-th particle.
In SPSO, particle $i$ updates its position by
\begin{equation}
  X^{j,k+1}_{i} =
  \left\{
  \begin{array}{lr}
   \{ e ~|~d(e \in K^{j}_{i} ~\mathrm{and} ~e \in \Omega^{j}) \} & (K^{j}_{i} \in \Omega^{j}) \\
   \{ e ~|~d(e \in X^{j,k}_{i} ~\mathrm{and} ~e \in \Omega^{j}) \} & (K^{j}_{i} \notin \Omega^{j}_{i} ~\mathrm{and} ~X^{j}_{i} \in \Omega^{j}) \\
   \{ e ~|~d(e \in \Omega^{j}) \} & (\mathrm{otherwise}) 
  \end{array}
  \right.
\end{equation}
where $\Omega^{j}$ is the constraints, the list of the current selectable cities to avoid the previous visiting cities once.
$d(*)$ has the shortest distance between cities in the set.

For example, if particle $i$ starts from city 1, $K^{1}_{i} = \{(1,3),(1,6)\}$ is given in [12]. 
Constraints $\Omega^{j}$ is a set of edges to the next selectable cities. In this case, we have $\Omega^{1} = \{(1,2),(1,3),(1,4),$ $(1,5),(1,6)\}$.
In Eq.17, since the first condition $(K^{1}_{i} \in \Omega{1})$ is satisfied, Edge function $d(e \in K^{1}_{i} ~\mathrm{and} ~e \in \Omega^{1})$ returns edge $(1,6)$ which is the shortest in the set.
Therefore, we obtain $X^{1}_{i} = \{(1,6)\}$ and construct the solution.

\section{Set-Based Comprehensive Learning and Particle Swarm Optimization with Memory} 
In this section, we describe an algorithm of set-based comprehensive learning and particle swarm optimization with memory, 
which is an approach of searching for the best position in a set by referring the current position when updating the solution.
At the beginning, we redefine $K^{\prime}$ without the current position in the following set:
\begin{equation}
  {K^{\prime}}^{j}_{i} = \{ e~|~e \in K^{j}_{i} ~\mathrm{and} ~e \notin X^{j}_{i} \}
\end{equation}

This is the same as $ K - X $, namely $K^{\prime} = K - X$, which consists only of elements that may be improved.
The updating rule of position is denoted as follows:
\begin{equation}
  X^{j,k+1}_{i} =
  \left\{
  \begin{array}{lr}
   \{ e ~|~d(e \in {K^{\prime}}^{j}_{i} ~\mathrm{and} ~e \in \Omega^{j}) \} & ({K^{\prime}}^{j}_{i} \in \Omega^{j}) \\
   \{ e ~|~d(e \in X^{j,k}_{i} ~\mathrm{and} ~e \in \Omega^{j}) \} & ({K^{\prime}}^{j}_{i} \notin \Omega^{j}_{i} ~\mathrm{and} ~X^{j,k}_{i} \in \Omega^{j}) \\
   \{ e ~|~d(e \in \Omega^{j}) \} & (\mathrm{otherwise}) 
  \end{array}
  \right.
\end{equation}
SCLPSO learns the elements in $K$ at first and does the elements in $X$ at the rest.
As there is much learning of the elements in $X$ when there is little learning in $K$,
SCLPSO is difficult to update the solution.
On the other hand, SCLPSOM evaluates whether a candidate solution exists or not in order of $K^{\prime}$, $X$, and $E$ for each dimension.
SCLPSOM conducts the solution in the set satisfied constraints by repeating the operation to generate the solution with function $d(*)$.
In case of no candidate solution in $K^{\prime}$ and $X$, the solution is searched by using $E$.
As SCLPSOM preferentially learns the set with high possibility, it restores the solution in $K^{\prime}$. 
The position updating procedure that represents in pseudocode is shown in Fig.1.

\begin{figure}[htb]
  \begin{algorithmic}[1]
    \Procedure{$\mathrm{position\_updating}(X_{i}, V_{i})$}{}
    \State generate a random value $\alpha_{i} \in (0.0,1.0)$;
    \For{each dimension $j~(j=1,2,...,M)$}
      \State $K^{j}_{i} = \{ e~|~e / p(e) \in V^{j}_{i} ~\mathrm{and} ~ p(e) \geq \alpha_{i} \}$;
    \EndFor
    \For{each dimension $j~(j=1,2,...,M)$}
      \State ${K^{\prime}}^{j}_{i} = \{ e~|~e \in K^{j}_{i} ~\mathrm{and} ~e \notin X^{j}_{i} \}$;
    \EndFor
    \State ${NEW\_X}^{j}_{i} = \phi$;
    \For{each dimension$j~(j=1,2,...,M)$}
      \State ${Candidate\_Set}^{j}_{i} = \{e~|~e \in {K^{\prime}}^{j}_{i} ~\mathrm{and} ~e ~\mathrm{satisfies} ~\Omega^{j} \}$;
      \If{${Candidate\_Set}^{j}_{i} \neq \phi $}
        \State select an element from ${Candidate\_Set}^{j}_{i}$ and add it to ${NEW\_X}^{j}_{i}$;
      \Else
        \State ${Candidate\_Set}^{j}_{i} = \{e~|~e \in X^{j}_{i} ~\mathrm{and} ~e ~\mathrm{satisfies} ~\Omega^{j} \}$;    
        \If{${Candidate\_Set}^{j}_{i} \neq \phi $}
          \State select an element from ${Candidate\_Set}^{j}_{i}$ and add it to ${NEW\_X}^{j}_{i}$;
        \Else
          \State ${Candidate\_Set}^{j}_{i} = \{e~|~e \in E^{j} ~\mathrm{and} ~e ~\mathrm{satisfies} ~\Omega^{j} \}$;
          \State select an element from ${Candidate\_Set}^{j}_{i}$ to repair ${NEW\_X}^{j}_{i}$;
        \EndIf      
      \EndIf
    \EndFor
    \State $X_{i} = {NEW\_X}_{i}$;
    \EndProcedure
  \end{algorithmic}
  \caption{Pseudocode for the position updating procedure.}
  \label{fig:1}
\end{figure}

After generating $K^{j}_{i}$ with possibilities, we create ${K^{\prime}}^{j}_{i}$ with elements of $K^{j}_{i}$ that are subtracted elements of the previous $X^{j}_{i}$,
where $M$ is the number of cities (line 2 to 8 in Fig.1).
For each dimension $j$, particle $i$ first learns from the element in ${K^{\prime}}^{j}_{i}$ and adds it to $NEW\_X^{j}_{i}$ (line 11 to 13 in Fig.1).
If there is no available element in the set, particle $i$ uses the element in the other set to build $NEW\_X^{j}_{i}$ (line 14 to 21 in Fig.1).
When select an element from the candidate set, using the edge function (line 13, 17 and 20 in Fig.1).
After all $NEW\_X^{j}_{i}$ are completed, the construction of $NEW\_X^{j}_{i}$ terminates.

\section{Numerical Experiments}
We examine numerical experiments for simple genetic algorithm (SGA), max-min ant system (MMAS), DPSO, MBPSO, SCLPSO, and our algorithm(SCLPSOM).
TSP instances are eil51, berlin52, st70, eil76, rat99, kroA100, eil101, ch130, kroA150, and d198 in TSPLIB [10].
We give parameters: the number of city $n$, population size $N = 20$, number of maximum iteration $I_{max} = 10000 n / N$, and run times $R = 30$ for each algorithm.
Other parameters are chosen as follows: crossover rate 0.6 in SGA, 
$\alpha = 1$, $\beta = 2$, and evaporation coefficient is $\rho = 0.98$ in MMAS,
inertia weight $\omega = 0.9 - 0.5 * k / I_{max}$ and weighting constant $c = 2.0$ in PSO, and
probability generated by a uniform random value $r$ in $[0.0,1.0]$.

\begin{table}[htb]
  \centering
  \caption{Experimental results.}
  \label{table:1}
  \scalebox{0.85}{
    \begin{tabular}{c|c|c|c|c|c|c|c|c}
      \hline
      Instance & Best Known & & SGA & MMAS & DPSO & MBPSO & SCLPSO & SCLPSOM \\
      \hline
      
      eil51 & 426 & Best & 470 & \textbf{426} & \textbf{426} & \textbf{426} & \textbf{426} & \textbf{426} \\  
            &     & Worst & 593 & 430 & 438 & 428 & \textbf{427} & \textbf{427} \\ 
            &     & Mean & 514.433 & 427.667 & 431.367 & 427.267 & 426.600 & \textbf{426.200} \\ 

      \hline
      berlin52 & 7542 & Best & 7758 & \textbf{7542} & \textbf{7542} & \textbf{7542} & \textbf{7542} & \textbf{7542} \\ 
              &      & Worst & 10405 & \textbf{7542} & 7961 & \textbf{7542} & \textbf{7542} & \textbf{7542} \\ 
              &      & Mean & 8884.800 & \textbf{7542.000} & 7648.167 & \textbf{7542.000} & \textbf{7542.000} & \textbf{7542.000}\\ 

      \hline
      st70 & 675 & Best & 728 & \textbf{675} & \textbf{675} & \textbf{675} & \textbf{675} & \textbf{675} \\ 
          &     & Worst & 974 & 700 & 700 & 686 & 677 & \textbf{675} \\ 
          &     & Mean & 832.900 & 683.367 & 686.133 & 678.267 & 675.333 & \textbf{675.000}\\ 

      \hline
      eil76 & 538 & Best & 579 & \textbf{538} & \textbf{538} & \textbf{538} & \textbf{538} & \textbf{538} \\ 
            &     & Worst & 833 & 545 & 559 & 547 & 541 & \textbf{538} \\ 
            &     & Mean & 657.267 & 540.033 & 547.033 & 539.433 & 538.333 & \textbf{538.000}\\ 
      \hline
      rat99 & 1211 & Best & 1358 & \textbf{1211} & 1217 & \textbf{1211} & \textbf{1211} & \textbf{1211}\\
            &      & Worst & 1984 & 1224 & 1279 & 1234 & 1213 & \textbf{1212}\\
            &      & Mean & 1563.300 & 1213.933 & 1242.033 & 1218.733 & 1211.567 & \textbf{1211.033}\\
      \hline
      kroA100 & 21282 & Best & 23355 & \textbf{21282} & \textbf{21282} & \textbf{21282} & \textbf{21282} & \textbf{21282} \\ 
              &       & Worst & 30237 & 21762 & 21865 & 21396 & \textbf{21282} & \textbf{21282} \\ 
              &       & Mean & 26505.330 & 21421.87 & 21510.030 & 21338.430 & \textbf{21282.000} & \textbf{21282.000}\\ 
      \hline
      eil101 & 629 & Best & 727 & 630 & 634 & \textbf{629} & \textbf{629} & \textbf{629} \\ 
            &     & Worst & 987 & 650 & 657 & 641 & 642 & \textbf{630} \\ 
            &     & Mean & 824.300 & 640.300 & 643.633 & 634.400 & 631.067 & \textbf{629.367}\\ 

      \hline
      ch130 & 6110 & Best & 7247 & 6160 & 6164 & \textbf{6110} & 6157 & \textbf{6110} \\
            &      & Worst & 8914 & 6262 & 6328 & 6297 & 6308 & \textbf{6159} \\ 
            &      & Mean & 8208.400 & 6199.533 & 6230.467 & 6189.947 & 6216.467 & \textbf{6126.767}\\

      \hline
      kroA150 & 26524 & Best & 33040 & 26696 & 26580 & 26564 & \textbf{26524} & \textbf{26524} \\ 
              &       & Worst & 41294 & 27248 & 27467 & 27331 & 27108 & \textbf{26709} \\ 
              &       & Mean & 36718.030 & 27015.070 & 27136.130 & 26938.770 & 26852.400 & \textbf{26567.900}\\ 

      \hline
      d198 & 15780 & Best & 17790 & 15837 & 19852 & 15873 & 16350 & \textbf{15780} \\
          &       & Worst & 20028 & 16065 & 20913 & 16756 & 17427 & \textbf{15844} \\ 
          &       & Mean & 18770.270 & 15891.170 & 20632.000 & 16208.800 & 16810.970 & \textbf{15797.370}\\ 

      \hline
    \end{tabular}
  }
\end{table}

Results are given in Table I. In the instances with a small number of cities for kroA100, the fitness difference among the existing algorithms except SGA and SCLPSOM is not large.
As the number of cities increases, the difference among the averages of SCLPSOM and other algorithms becomes larger.
SCLPSOM is only able to find the optimal solution for all instances.

\begin{table}[htb]
  \centering
  \caption{Comparison results of SCLPSOM with other algorithm.}
  \label{table:2}
  \begin{tabular}{c|ccc}
    \hline
    Algorithms& $+$ & $=$ & $-$\\ \hline
    SGA & 10 & 0 & 0\\ \hline
    MMAS & 9 & 1 & 0\\ \hline
    DPSO & 10 & 0 & 0\\ \hline
    MBPSO & 9 & 1 & 0\\ \hline
    SCLPSO & 6 & 4 & 0\\ \hline
  \end{tabular}   
\end{table}

\begin{table}[htb]
	\centering
	\caption{Results of computation time (ms).}
	\label{table:3}
	\scalebox{0.85}{
		\begin{tabular}{c|c|c|c|c|c|c}
			\hline
			Instance & SGA & MMAS & DPSO & MBPSO & SCLPSO & SCLPSOM\\ \hline 
			kroA100 & 118136 & 201594 & 138715 & 123731 & 84078 & 88399 \\ \hline
			kroA150 & 563355 & 677338 & 452006 & 416224 & 291277 & 305337\\ \hline
		\end{tabular}
	}
\end{table}

Wilcoxon signed rank tests are conducted between SCLPSOM and
other algorithm for each instances given in Table II.
Three symbols "+, -, =" indicate that SCLPSOM exhibits significantly better, worse than, and equal to the competitor, respectively.
Table II shows that there are significant differences in all cases for our algorithm.

Results of the computation time for all algorithms for kroA100 and kroA150 are shown in Table III.
The computation time of SCLPSO and SCLPSOM are faster than that of the existing algorithms. 
The computational complexity of generating a solution in SCLPSO and SCLPSOM is $O(n^{2})$.

For spec of a computer, we performed on a machine with Intel(R) Core(TM) i9-10900 at 2.80GHz and 32.0GB of RAM.
The operating system is Windows 10 Enterprise 64 bit.
Programming language is Java. Compiler and its version are JAVA SE-11.\\

\section{Conclusions}
We have present a novel algorithm and set-based comprehensive learning and particle swarm optimization with memory (SCLPSOM) for discrete optimization problem.
This approach was to search for the best position in a set by referring the current position when updating the solution.
SCLPSOM showed the effectiveness compare to the existing algorithms.
The future work is theoretical considerations.


\begin{thebibliography}{99}

\bibitem{REF Introduction}
E. Aiyoshi, and K. Yasuda,
\textit{Metaheuristics and Applications},
Ohmsha, Japan, 2007.

\bibitem{REF Introduction}
D.E. Goldberg,
\textit{Genetic Algorithms in Search, Optimization, and Machine Learning},
AddisonWesley Longman Publishing Co., Inc, 1989.

\bibitem{REF ACO}
M. Dorigo, V. Maniezzo, and A. Colorni, 
``Any system: Optimization by a colony of cooperating agents,"
\textit{IEEE Transactions on System,Man, and Cybernetics}, part B, 26, 1, pp.\ 29--41, 1996.

\bibitem{REF PSO}
J. Kennedy, and R.C. Eberhart, 
``Particle swarm optimization,"
\textit{Proceedings of the IEEE International Conference on Neural Networks}, pp.\ 1942--1948, 1995.

\bibitem{REF DPSO}
J. Kennedy, and R.C. Eberhart, 
``A discrete binary version of the particle swarm algorithm,"
\textit{Proceeding of the IEEE International Conference on Systems Man and Cybernetics}, pp.\ 4104--4109, 1997.

\bibitem{REF MBPSO}
Z. ji, T. Tian, and S. He, 
``A memory binary particle swarm optimization," 
\textit{Proceedings of the IEEE International Conference on Evolutionary Computation}, pp.\ 1--5, 2012.

\bibitem{REF SPSO}
W.N. Chen, J. Zhang, and H.S.H. Chung, 
``A novel set-based particle swarm optimization method for discrete optimization problems,"
\textit{IEEE Transactions on Evolutionary Computation}, vol.\ 14, no.\ 2, pp.\ 278--299, 2010.

\bibitem{REF MMAS}
T. Stuzle, and H. Hoos, 
``Max-min ant system," 
\textit{Generation Computer Systems}, vol.\ 16, no.\ 9, pp.\ 889--914, 2004.

\bibitem{REF CLPSO}
J.J. Liang, A.K. Qin, and P. Suganthan, 
``Comprehensive learning particle swarm optimizer for global optimization of multimodal functions,"
\textit{IEEE Transactions on Evolutionary Computation}, vol.\ 10, no.\ 3, pp.\ 281--295, 2006.

\bibitem{REF TSP}
TSPLIB \url{http://comopt.ifi.uni-heidelberg.de/software/TSPLIB95/}.

\bibitem{REF SPSOSM}
T. Hino, S. Ito, T. Liu, and M. Maeda, 
``Set-based particle swarm optimization with status memory for knapsack problem,"
\textit{Artificial Life and Robotics}, vol.\ 21, no.\ 1, pp.\ 98--105, 2016.

\bibitem{REF SPSOMS}
M. Maeda, and T. Hino, 
``A novel approach of set-based particle swarm optimization with memory state,"
\textit{MIWAI 2017, LNAI10607}, pp.\ 437--449, 2017.

\end{thebibliography}
\end{document}